\documentclass[11pt]{article}

\usepackage[final]{acl}

\usepackage{times}
\usepackage{latexsym}

\usepackage[T1]{fontenc}

\usepackage[utf8]{inputenc}

\usepackage{microtype}

\usepackage{inconsolata}

\usepackage{graphicx}
\usepackage{lipsum}
\usepackage{hyperref}
\usepackage{booktabs}
\usepackage{dsfont}
\usepackage{enumitem}
\usepackage{amsmath}
\usepackage{amsfonts}
\usepackage{xcolor}
\usepackage{tabularx}
\usepackage{multirow}
\usepackage{makecell}   
\usepackage{booktabs}   
\usepackage{pifont}     
\usepackage{mfirstuc}
\usepackage{tikz}
\usepackage{longtable}
\usepackage{array}
\usepackage{subcaption}
\usetikzlibrary{positioning, calc,matrix,fit,backgrounds}

\usepackage{cleveref}

\makeatletter
\newcommand{\reviewdependent}[2]{%
  \ifacl@finalcopy
    #2%
  \else
    #1%
  \fi
}
\makeatother

\definecolor{greenfill}{HTML}{D5E8D4}
\definecolor{greenborder}{HTML}{82B366}
\definecolor{greentext}{HTML}{556B43 } 

\definecolor{bluefill}{HTML}{DAE8FC}
\definecolor{blueborder}{HTML}{6C8EBF}
\definecolor{bluetext}{HTML}{4C588E}

\definecolor{redfill}{HTML}{F8CECC}
\definecolor{redborder}{HTML}{B85450}
\definecolor{redtext}{HTML}{843C39}

\definecolor{lilafill}{HTML}{E1D5E7}
\definecolor{lilaborder}{HTML}{9673A6}
\definecolor{lilatext}{HTML}{60496A} 

\newcommand{\cmark}{\ding{51}}
\newcommand{\xmark}{\ding{55}}
\newcommand{\piname}{expected training contribution}

\newcommand{\unconstrained}{unconstrained}
\newcommand{\Unconstrained}{Unconstrained}

\newcommand{\noretention}{no retention}

\newcommand{\NoRetention}{No Retention}

\newcommand{\noreentry}{no reentrance}

\newcommand{\NoReentry}{No Reentrance}

\newcommand{\fullretention}{full retention}

\newcommand{\FullRetention}{Full Retention}

\newcommand{\universala}{universal}

\newcommand{\UniversalA}{Universal}

\newcommand{\numpapers}{90}
\newcommand{\numstrategies}{107}

\title{Disentangling Curriculum Learning in NLP:\\Towards a Unifying Taxonomy}

\author{
  Vanessa Toborek$^{1,2}$,
  Florian Seiffarth$^{1,2}$,
  Sebastian M\"uller$^{1,2}$,
  \and
  \textbf{Tam\'as Horv\'ath}$^{1,2,3}$
  \\
  $^1$University of Bonn, Germany,
  $^2$Lamarr Institute, Germany,
  $^3$Fraunhofer IAIS, Germany
  \\
  \texttt{vtoborek@uni-bonn.de}
}

\date{}

\begin{document}
\maketitle
\begin{abstract}
Despite more than a decade of curriculum learning (CL) research in NLP, the field lacks a principled account of which difficulty function or scheduler to use for a given problem. 
To understand what has hindered progress towards this account, we propose a fine-grained taxonomy separating difficulty evaluation from training scheduling to enable systematic analysis of CL strategies. 
For difficulty evaluation, we distinguish \textit{attribution source} and \textit{task dependence}, revealing difficulty as a perspectival concept encoding different assumptions about what makes an instance hard to learn.
For scheduling, we provide the first formalisation of CL schedulers in terms of \textit{expected training contribution}, enabling comparison across implementations by introducing \textit{retention regimes} and \textit{monotonicity properties}.
Applied in the first dedicated survey of CL in NLP, covering \numstrategies{} strategies from \numpapers{} papers, our taxonomy reveals a systematic \textit{incomparability problem}: prior works conflate distinct notions of difficulty and scheduling, often pursuing different objectives under the same CL label -- hindering comparison and the accumulation of a coherent evidence base. 
Beyond diagnosis, the taxonomy supports the design, analysis, and comparison of CL strategies, and motivates evaluation practices that disentangle the sources of observed improvement.
\end{abstract}

\section{Introduction}
\label{sec:intro}

Data-level curriculum learning (CL) is a long-standing approach in machine learning that aims to improve training by controlling the exposure of training instances to a model \citep{Elman1993,Bengio2009}. Inspired by human learning, the core intuition is that models may benefit from starting training with ``easier'' instances and moving to ``harder'' ones, thereby improving convergence, generalisation, or optimisation stability. Any CL strategy therefore requires two components: a \textit{difficulty function} to evaluate training instances and a \textit{scheduler} to realise the progression from easy to hard \cite{Wang2021a,Soviany2022}. 

Despite sustained interest in CL, including benchmark-level comparisons across methods, datasets, and model families \citep{Wu2021WhenDoCurriculaWork,Zhou2024CurBench}, the field has not converged on which difficulty function or scheduler to use for a given problem. 
We argue that this is not due to a lack of empirical work, but to the absence of a shared framework precise enough to establish what a CL strategy implements.
Without such a framework, 
findings reported under the same CL label may concern structurally different -- and thus incomparable -- strategies, preventing clear attributable analysis.

The lack of conceptual specification is particularly pronounced for the definition of difficulty in NLP. 
Existing curricula draw on heterogeneous notions of difficulty from adjacent research areas like readability assessment \cite{pucci-etal-2023-languages}, text simplification \cite{agrawal-carpuat-2022-imitation}, psycholinguistics \cite{tsvetkov-etal-2016-learning}, and annotation studies \citep{elgaar-amiri-2023-hucurl}. Model-centric CL approaches further amplify this heterogeneity by defining difficulty through training dynamics rather than input properties \citep{maharana-bansal-2022-curriculum}.
These difficulty functions are not interchangeable: they differ in whose difficulty they capture, what information they require, and how transferable they are across settings.

To address this, we propose a two-part taxonomy of CL that analytically separates difficulty evaluation from training scheduling.
On the difficulty side, we build on \citet{Toborek2026} and characterise difficulty functions by \textit{attribution source} and \textit{task dependence}, yielding four quadrants that expose their assumptions and information requirements.
On the scheduling side, we formalise schedulers in terms of the \textit{expected training contribution} they realise and introduce two mechanism-independent axes: \textit{data-retention regimes} and three \textit{monotonicity properties}.
This unified view enables direct comparison of sampling- and weighting-based schedulers for the first time and allows us to analyse their induced training progression.

The taxonomy reveals two aspects of an \textit{incomparability problem} in CL: difficulty functions encode distinct assumptions about what makes an instance easy to learn, making results hard to attribute to a shared mechanism, while schedulers implementing an easy-to-hard progression need not be monotone in the same sense nor expose the model to comparable training distributions. Applying this taxonomy to \numstrategies{} from \numpapers{} works from the ACL Anthology further reveals another source of incomparability: many curricula pursue objectives beyond pure difficulty-based ordering, such as data quality or domain filtering, by assuming that some instances are less useful for training.

By disentangling previously conflated concepts in CL, our taxonomy fosters comparability and the structured development of novel approaches.

In summary, our main contributions are:
\setlist{nolistsep}
\begin{itemize}[noitemsep]
    \item We introduce a two-part taxonomy separately analysing difficulty evaluation and training scheduling in data-level CL (\Cref{sec:taxonomy}).

    \item We use this taxonomy to present the first dedicated survey of CL in NLP, analysing \numstrategies{} papers by difficulty quadrants, retention regimes, and monotonicity properties (\Cref{sec:survey}).

    \item We identify sources of incomparability in CL for NLP and recommend better practices for reporting and evaluation (\Cref{sec:discussion}).
\end{itemize}

We publish the annotated set of strategies as an interactive online resource at \url{https://vtoborek.github.io/unifying-taxonomy-cl-nlp/}, open to contributions from the community.

\section{Background on Curriculum Learning}
\label{sec:background}

We briefly introduce the terminology needed to position our taxonomy. 
Existing surveys offer comprehensive introductions to CL and broad overviews of the CL literature, including works in NLP \citep{Wang2021a,Soviany2022}.
However, they primarily organise methods according to implementation paradigms or application domains.
In contrast, our taxonomy provides a common representation for comparing CL strategies across implementations based on their underlying difficulty and scheduling properties.

\medskip
\noindent \textbf{Curriculum Learning as a Two-Part Strategy.}
Any CL strategy can be decomposed into two parts \citep{Wang2021a,Soviany2022}. A \textit{difficulty function} assigns a notion of difficulty to training instances, while a \textit{training scheduler} determines how this difficulty influences training over time, e.g. by controlling sampling probabilities or loss weighting. Throughout this survey, we use the term scheduler as a unifying abstraction for all mechanisms that govern how training data is exposed over time; pacing functions (cf.\ \citet{Hacohen2019}) are treated as particular parametrisations of a scheduler. In the formulation of \citet{Mitchell1997} both operate on training data $E$, where a model $M$ learns task $T$ with respect to some performance measure $P$ from experience $E$. 

\medskip
\noindent \textbf{Data- vs.\ Task-Level CL.}
We distinguish \textit{data-level} CL, which reorders instances under a fixed target distribution $p(y \mid x)$, from \textit{task-level} CL, where training stages involve incompatible target relations. For instance, transitioning from binary sentiment to fine-grained emotion labels is task-level CL, as no single $p(y \mid x)$ reconciles both label schemes. Data-level CL is the more widely adopted variant \citep{Soviany2022}.

\medskip
\noindent \textbf{Modes of Difficulty Evaluation.}
We further distinguish three modes of difficulty evaluation. 
\textit{Global} evaluation induces a static ordering or partitioning of the dataset by assigning each training instance a single value, computed once, typically prior to training, and fixed throughout. 
\textit{Local} evaluation instead depends on the current model state, e.g. model parameters, loss, or confidence. Hence, the difficulty of a given sample may change in the course of training. 
Finally, \textit{in-sample} evaluation associates multiple values with a single instance by actively modifying it. 
Typical approaches gradually increase input length or adjust perturbation strength, defining difficulty over transformed variants rather than fixed samples.

\section{Taxonomy}
\label{sec:taxonomy}

\begin{figure}
    \centering
    \begin{tikzpicture}[
    outerbox/.style={
        rectangle,
        rounded corners=5pt,
        draw=black,
        text=bluetext,
        line width=0.5pt,
        inner sep=5pt,
        fill=none
    },
    titletext/.style={
        font=\fontsize{10pt}{10pt}\selectfont\bfseries,
        text=blueborder,
        align=center
    },
    boxbase/.style={
        rectangle,
        rounded corners=4pt,
        align=center,
        inner sep=3pt,
        minimum height=0.65cm,
        font=\fontsize{8.2pt}{8.8pt}\selectfont,
        line width=0.5pt
    },
    greenbox/.style={
        boxbase,
        fill=greenfill,
        draw=greenborder,
        text=greentext,
        fill opacity=0.7,
        draw opacity=0.7,
        text opacity=1
    },
    lilabox/.style={
        boxbase,
        fill=lilafill,
        draw=lilaborder,
        text=lilatext,
        fill opacity=0.7,
        draw opacity=0.7,
        text opacity=1
    },
    listcell/.style={
        align=left,
        anchor=north,
        font=\fontsize{7pt}{8pt}\selectfont,
        text=black,
        inner sep=1pt,
        minimum width=1.45cm
    }
]

\node[titletext] (title) at (0,0) {Curriculum Learning};

\node[
    greenbox,
    below=0.20cm of title,
    minimum width=4.75cm
] (df) {
    Difficulty Function\\[-1pt]
    {\fontsize{7pt}{8pt}\selectfont(§\labelcref{sec:taxonomy-quadrants})}
};

\matrix (dm) [
    matrix,
    below=0.13cm of df,
    column sep=0.25cm,
    row sep=0.10cm
] {
    \node[greenbox, minimum width=2.25cm, minimum height=0.78cm] (d1)
        {attribution-\\[-1pt]source}; &
    \node[greenbox, minimum width=2.25cm, minimum height=0.78cm] (d2)
        {task-\\[-1pt]dependence}; \\

    \node[listcell] (d1list)
        {$\circ$~human-centric\\
         $\circ$~model-centric}; &
    \node[listcell] (d2list)
        {$\circ$~agnostic\\
         $\circ$~dependent}; \\
};

\node[
    lilabox,
    below=0.20cm of dm,
    minimum width=4.75cm
] (ts) {
    Training Scheduler\\[-1pt]
    {\fontsize{7pt}{8pt}\selectfont(§\labelcref{sec:taxonomy-schedulers})}
};

\matrix (sm) [
    matrix,
    below=0.13cm of ts,
    column sep=0.13cm,
    row sep=0.10cm
] {


    \node[lilabox, minimum width=2.25cm, minimum height=0.78cm] (s1)
        {retention regime\\[-1pt]
        {\fontsize{6.2pt}{6.8pt}\selectfont(§\labelcref{sec:scheduler-retention})}}; &
    \node[lilabox, minimum width=2.25cm, minimum height=0.78cm] (s2)
        {monotonicity\\[-1pt]
        {\fontsize{6.2pt}{6.8pt}\selectfont(§\labelcref{sec:scheduler-monotonicities})}}; \\
        
    \node[listcell] (s1list)
        {$\circ$~\noretention{}\\
         $\circ$~\noreentry{}\\
         $\circ$~\fullretention{}\\
         $\circ$~\universala{}\\
         $\circ$~\unconstrained{}}; &


    \node[listcell] (s2list)
        {$\circ$~selection\\
         $\circ$~frontier\\
         $\circ$~exposure}; \\
};

\begin{scope}[on background layer]
\node[
    outerbox,
    fit=(title)(df)(dm)(ts)(sm),
    inner xsep=0.14cm,
    inner ysep=0.16cm
] {};
\end{scope}

\end{tikzpicture}
    \caption{Overview of our proposed dual taxonomy concerning difficulty evaluation and training scheduling.}
    \label{fig:taxonomy-overview}
\end{figure}
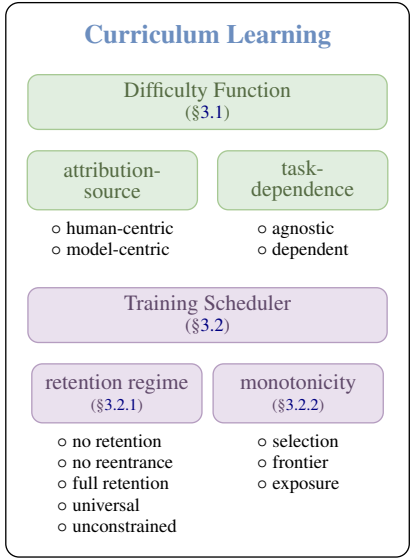

Curriculum learning still lacks a shared formalism: difficulty is typically defined ad hoc, schedulers are described informally, and strategies accumulate without a principled basis for comparison.

We address this gap with a \textit{dual taxonomy} that treats difficulty evaluation (§~\labelcref{sec:taxonomy-quadrants}) and training scheduling (§~\labelcref{sec:taxonomy-schedulers}) as distinct but complementary parts of curriculum design.
\Cref{fig:taxonomy-overview} provides an overview of all components of the taxonomy.

\subsection{Difficulty Functions}
\label{sec:taxonomy-quadrants}

The space of difficulty functions in CL is highly heterogeneous, especially in NLP, where curricula draw on proxies ranging from surface input features and readability measures to annotation behaviour, auxiliary models, and target-model training dynamics. One way to organise this space is by the information used to estimate instance difficulty. 

Let $d_t : \mathcal{D} \rightarrow \mathbb{R}$ be a difficulty function assigning a scalar value $d_t(x)$ to each training instance $x$ at stage $t$.
Difficulty functions may rely on auxiliary resources $R$ required to compute these values.
In this case, we make this dependency explicit by writing $d^R_t(x)$.
When the auxiliary resources are not relevant to the discussion, we simply write $d_t$.
Under global evaluation as defined in \Cref{sec:background}, the difficulty function $d_t$ is fixed across stages, whereas under local evaluation it may change with the model.\footnote{In-sample evaluation can be reduced to global evaluation by expanding $\mathcal{D}$ to include all generated variants as distinct elements, so it does not require separate discussion.}

The diversity of possible resources makes an exhaustive categorisation neither practical nor particularly informative.
We adopt the categorisation proposed by \citet{Toborek2026} and organise difficulty functions along two orthogonal dimensions of their information requirements.
Let $R_T$ denote resources defined with respect to the target task and $R_M$ resources obtained from one or more models.
A difficulty function is \textit{task-dependent} if $R\cap R_T\neq\emptyset$ and \textit{task-agnostic} otherwise; it is \textit{model-centric} if $R\cap R_M\neq\emptyset$ and \textit{human-centric} otherwise.
Task-specific resources include not only target labels, but any information whose interpretation depends on the prediction objective, such as reference outputs, annotation agreement, or task-specific model signals.

Practically, classifying a difficulty function reduces to two questions:
\begin{enumerate}[noitemsep]
    \item Does computing difficulty require information specific to the target task?
    \item Does computing difficulty require information obtained from one or more models?
\end{enumerate}

Together, these dimensions define four difficulty quadrants, which organise our survey of difficulty proxies:

\begin{description}[noitemsep,labelindent=0.35cm,leftmargin=1.85cm,labelwidth=1.2cm,labelsep=0.3cm]
    \item[TA-H:] $R \cap R^T = \emptyset \land R \cap R^M = \emptyset$ (task-agnostic, human-centric), including, e.g., readability metrics and surface features;
    \item[TA-M:] 
    $R \cap R^T = \emptyset \land R \cap R^M \neq \emptyset$ 
    (task-agnostic, model-centric), e.g. perplexity and embedding-based metrics;
    \item[TD-H:] 
    $R \cap R^T \neq \emptyset \land R \cap R^M = \emptyset$ 
    (task-dependent, human-centric), 
    e.g. label entropy and human-reference metrics;
    \item[TD-M:] 
    $R \cap R^T \neq \emptyset \land R \cap R^M \neq \emptyset$ 
    (task-dependent, model-centric), 
    e.g. loss and teacher-student signals.
\end{description}

These axes describe technical requirements, but also reflect assumptions about why an ordering should benefit training: for example, whether difficulty is expected to follow human-oriented linguistic complexity, model uncertainty, or task-specific errors. Results across quadrants therefore require care when compared, since they may not reflect alternative proxies for the same notion of difficulty.

\subsection{Training Scheduler}
\label{sec:taxonomy-schedulers}
A scheduler determines how difficulty evaluations shape training over time. 
While ``easy-to-hard'' captures the governing intuition, it leaves the exact structure of progression underspecified. 

Prior categorisations distinguish deterministic from probabilistic schedulers \citep{zhang-etal-2019-curriculum}, discrete from continuous ones \citep{christopoulou-etal-2022-training}, or pre-defined from automatic pacing \citep{Wang2021a} describing implementation choices rather than how training focus progresses over time.

We address this gap with a formal framework centred on the \textit{\piname{}} $\pi_t$, a unified abstraction of all scheduling choices. The framework yields two axes of mechanism-agnostic comparison: \textit{retention regimes} and \textit{monotonicity properties}.
This framework enables us to (i) test easy-to-hard exposure in detail, (ii) identify functionally equivalent schedulers, and (iii) reveal an attribution problem under local difficulty evaluation, where monotone progression may stem from either scheduling or evolving difficulty estimates.

\medskip
\noindent \textbf{Setup.}
A \textit{scheduler} is a mapping $\mathcal{S}:(d_t, t) \mapsto \pi_t$, where $\pi_t \in \Delta(\mathcal{D})$, $\Delta(\mathcal{D})$ denotes the set of all possible distributions over $\mathcal{D}$, and $d_t$ is a difficulty function as defined below. 
Intuitively, $\pi_t(x)$ encodes the expected contribution of instance $x \in \mathcal{D}$ to the parameter update at stage $t$.
Thus, we call $\pi_t$ the \textit{\piname{}} at stage $t$

\medskip
\noindent \textbf{Derived objects.} For stage $t \ge 1$, we define the \textit{admissible pool} as the support of $\pi_t$:
{\abovedisplayskip=3pt
\belowdisplayskip=3pt
\abovedisplayshortskip=2pt
\belowdisplayshortskip=2pt
\[
A_t := \operatorname{supp}(\pi_t) = \{x \in \mathcal{D} : \pi_t(x) > 0\}.
\]
}
\par\noindent $A_t$ reflects how schedulers are often designed in practice: rather than individually specifying the contribution of every instance, schedulers operate by deciding what enters or leaves the training pool. Instances outside $A_t$ cannot contribute to the parameter update at stage $t$.

To describe how $\pi_t$ evolves between consecutive stages, we further define for $t \ge 2$ the \textit{retained set} as $X_t := \{x\in\mathcal{D} : \pi_t(x)>0 \text{ and } \pi_{t-1}(x)>0\}$ and $C_t := \{x\in\mathcal{D} : \pi_t(x)>\pi_{t-1}(x)\}$ as the \textit{increase-set}. 
$X_t$ captures which instances retain positive contribution across consecutive stages;  $C_t$ identifies those whose contribution actively grows. Notably, $C_t$ is not restricted to newly admitted instances: it may also contain already-admissible samples whose contribution increases further. We deliberately track increases rather than decreases, because our goal is to characterise which instances can newly drive training at each stage.

\medskip
\subsubsection{Retention Regimes}
\label{sec:scheduler-retention}
A key axis along which schedulers in the literature diverge is their retention behaviour: how samples are retained, removed, or reintroduced in $A_t$.
By looking at the sequence $(\pi_t)_{t\geq 1}$, we are able to distinguish between an unconstrained baseline regime and a small set of recurring structured retention regimes, each capturing a characteristic pattern in the evolution of support independent of implementation.
This makes retention the first mechanism-agnostic axis of comparison in our framework.

\medskip
\noindent (1) \textit{\Unconstrained{} Regime.} No structural constraints are imposed on the evolution of $\pi_t$. Instances may enter, leave, and re-enter the support across stages without restriction. Serving as the most general reference class, this regime captures schedulers that generate arbitrary stage-wise slices.

\medskip
\noindent (2) \textit{\NoReentry{} Regime.} For all \(x\in\mathcal D\) and all \(t'\le t \le t''\),
$\pi_{t'}(x)>0 \;\wedge\; \pi_{t''}(x)>0 \;\implies\; \pi_{t}(x)>0.$
In this regime, previously exposed instances may be retained for an arbitrary number of stages, so $X_t$ may be non-empty. Likewise, $C_t$ may include both newly admitted and already-active instances. The only restriction is that once support for an instance becomes zero, it cannot become positive again, i.e. each instance contributes during at most one contiguous interval of stages.

\medskip
\noindent (3) \textit{\NoRetention{} Regime.} For all $t>1$ and $x\in\mathcal D$,
$\pi_t(x)>0 \;\implies\; \forall\, t'\neq t:\ \pi_{t'}(x)=0 .$
Each instance contributes during at most one stage, i.e. if a sample contributes at step $t$ it did not contribute in $t'<t$ and will not contribute in $t''>t$. As a consequence, $X_t=\emptyset$ and $C_t = A_t$ for all $t$. A canonical example of the no-retention regime is the One-Pass scheduler \cite{Bengio2009}, which trains on disjoint dataset partitions.

\medskip
\noindent (4) \textit{\FullRetention{} Regime.} 
For all $1 \leq t' < t$ and $x\in\mathcal D$,
$\pi_{t'}(x) > 0 \;\implies\; \pi_{t}(x)> 0.$
Retention is monotone at the level of support: once admitted, an instance cannot leave $A_t$, i.e. the interval of contribution extends until the end of training. Accordingly, $A_t$ can only expand, $X_t$ is never empty, as $\pi_t(x) > 0$ for at least one $x \in \mathcal{D}$, and $C_t$ may contain both newly admitted instances and already-admissible ones whose contribution increases further. Mass within the admissible pool may be redistributed freely. 
Baby-Step \cite{spitkovsky-etal-2010-baby} and competence-based schedulers \cite{platanios-etal-2019-competence} are canonical examples, progressively expanding the admitted set while renormalising contribution over $A_t$.
\medskip

\noindent (5) \textit{\UniversalA{} Regime.} For all $t\ge 1$ and $x\in\mathcal D$,
$\pi_t(x)>0 .$
All instances are admissible at every stage, so $A_t=\mathcal D$ for all $t$. Progression is expressed purely through changes in contribution magnitude rather than through admission or removal. Accordingly, $C_t$ isolates the subset whose contribution is currently increasing within the fixed full support.

\medskip
The five regimes defined in (1)--(5) above are partially ordered by inclusion, with:
{
\abovedisplayskip=3pt
\belowdisplayskip=3pt
\abovedisplayshortskip=2pt
\belowdisplayshortskip=2pt
\[ (5) \subset (4) \subset (2) \subset (1) \ \text{ and } \ (3) \subset (2) \enspace .\]}

In \Cref{sec:example-scheduler} of the appendix, we provide a worked example comparing two schedulers to illustrate how different retention regimes shape $\pi_t$.

\medskip
\noindent \textbf{Scheduler Implementation.}
Across the literature, the retention regimes above are realised by two mechanisms: \textit{sampling}, which defines a probability distribution $p_t$ over instances, and \textit{weighting}, which scales the contribution of each sampled instance to the loss by a factor $w_t$.
Since these mechanisms are not mutually exclusive and may be combined, we capture the joint effect in our framework as $\pi_t \propto p_t \times w_t$.
Setting $w_t(x)=1$ for all samples recovers pure sampling; keeping $p_t$ constant and varying only $w_t$ recovers pure weighting.

\medskip
\subsubsection{Monotonicity Properties} 
\label{sec:scheduler-monotonicities}
Retention regimes describe the structural evolution of $\pi_t$, but not whether a curriculum is actually easy-to-hard: a full-retention scheduler can admit harder or easier instances alike. To describe progression precisely, we introduce three monotonicity properties that track difficulty at different levels of aggregation, yielding a second mechanism-agnostic characterisation of any CL strategy. 

\medskip
\noindent (i) \textit{Selection Monotonicity.} Let \textit{selection difficulty} measure the difficulty of instances driving the transition from $\pi_t$ to $\pi_{t+1}$, given by the average instance difficulty in $C_t$:
$\mu_{\mathrm{sel}}(t) = \frac{1}{|C_t|}\sum_{x \in C_t} d_t(x)$.
A scheduler is \textit{selection-monotone} if $\mu_{\mathrm{sel}}(t+1)\ge \mu_{\mathrm{sel}}(t)$ for all $t$.

\medskip
\noindent (ii) \textit{Frontier Monotonicity.} 
Let $\tau(x) = \min \{t:x \in A_t\}$ denote the admission stage of $x$.
The \textit{difficulty frontier} measures the highest admission difficulty reached within $A_t$: $\iota(t) := \max_{x \in A_t} d_{\tau(x)}(x)$.
\footnote{ $d_{\tau(x)}(x)$ anchors the difficulty frontier to the scheduler decisions rather than the evolving model perception for local $d_t$; for global $d_t$, this reduces to $\iota(t) := \max_{x \in A_t} d_t(x)$.}
A scheduler is \textit{frontier-monotone} if $\iota(t+1) \geq \iota(t)$ for all $t$.

\medskip
\noindent (iii) \textit{Exposure Monotonicity.}
Let \textit{exposure difficulty} denote the expected difficulty under the \piname{}: $\mu_{\mathrm{exp}}(t) = \mathds{E}_{x \sim \pi_t} \big[d_t(x)\big]$.
A scheduler is \textit{exposure-monotone} if $\mu_{\mathrm{exp}}(t+1) \geq \mu_{\mathrm{exp}}(t)$ for all $t$. 

\medskip
We argue for the distinction between the three monotonicities, because progression enforced early need not be preserved downstream: a scheduler may admit increasingly harder instances without increasing the expected difficulty of the training signal, or may increase exposure through reweighting while leaving the frontier unchanged.

Formally, the three properties are defined over sequences of difficulty functions $(d_t)$, which may or may not remain stable across stages. Under global evaluation, monotonicity is directly attributable to the structural choices of the scheduler. Under local evaluation, $d_t$ co-evolves with the model: changes in $\mu_\mathrm{sel}$ or $\mu_\mathrm{exp}$ between stages may reflect scheduler decisions, model improvement, or both -- an \textit{attribution problem} our framework makes explicit. 
We therefore classify schedulers as monotone only when the property is \textit{structurally guaranteed}, i.e.\ holds regardless of how $d_t$ evolves.

\section{Survey of Curriculum Learning in NLP}
\label{sec:survey}
We survey the NLP literature in two steps: §~\labelcref{sec:survey-difficulty} reviews difficulty proxies through the four quadrants, and §~\labelcref{sec:survey-scheduler} reviews schedulers through the retention regimes. \Cref{sec:corpus-compilation} describes the corpus compilation and \Cref{tab:table-summary} in \Cref{sec:paper-analysis} presents the detailed paper-level analysis. 
\footnote{\reviewdependent{A full version of the paper-level analysis will be made available in a public repository upon acceptance, intended as an extensible resource to which future CL work may be added.}{We publish the paper-level analysis as an interactive community resource open to contributions: \url{https://vtoborek.github.io/unifying-taxonomy-cl-nlp/}.}}

\subsection{Survey of Difficulty Proxies}
\label{sec:survey-difficulty}

\begin{table*}
    \centering
    \fontsize{9pt}{8.5pt}\selectfont
    \begin{tabular}{p{1.2cm} p{4.8cm} p{7.5cm}}
\toprule
\textbf{Quadrant} & \textbf{Proxy Families} & \textbf{Examples}\\
\midrule

\multirow[c]{3}{*}{TA-H}
& psycholinguistic/ readability features & length, syntactic complexity, readability scores \\
& dataset-level attributes & genre, language similarity \\
& difficulty-by-construction & compositional depth, perturbations, context manipulation \\

\midrule

\multirow[c]{3}{*}{TA-M}
& auxiliary model signals & surprisal, cross-entropy\\
& pre-trained embedding geometry & vector norms, cosine/Euclidean distance \\
& LLM for evaluation/ data-generation & difficulty scores, data-level attributes\\

\midrule

\multirow[c]{4}{*}{TD-H}
& target complexity & output length, code/ graph/ proof depth \\
& target information injection & oracle pronouns and gold arguments \\
& human-reference discrepancy & edit distance, ROUGE, answerability \\
& direct human performance & annotation time and preference \\

\midrule

\multirow[c]{4}{*}{TD-M}
& current-model signals & confidence, loss, margin to the decision boundary \\
& auxiliary model signals & preference, relevance scores \\
& teacher-model signals & prediction confidence, question answering probability \\
& model-level perturbations & embedding perturbations, decision-boundary distance \\

\bottomrule
\end{tabular}
    \caption{Overview of difficulty-proxy families by quadrant.}
    \label{tab:overview-quad-survey}
\end{table*}

We survey difficulty proxies by quadrant as introduced in~\Cref{sec:taxonomy-quadrants}. \Cref{fig:counts} and \Cref{fig:mode-per-quadrant} in \Cref{sec:extended-quad-survey} show that proxies from TD-M are most common, TA-H and TA-M proxies are mostly global, and TD-M mostly local. \Cref{tab:overview-quad-survey} provides an overview of the different proxy families we identified within each quadrant; see \Cref{sec:extended-quad-survey} for an extended account of the literature with examples.

\medskip
\noindent \textbf{TA-H} proxies are the most accessible, because they rely on human-designed, label-free properties of the input, making them inexpensive to compute and often treated as default curricula. 
The dominant family of proxies relies on psycholinguistic and readability-related signals \cite{platanios-etal-2019-competence,tsvetkov-etal-2016-learning}. Other proxies employ source-level attributes \citep{elbayad-etal-2023-fixing,hong-etal-2025-game,elzohbi-zhao-2025-tahdib} or proxies defined by construction like targeted perturbations \cite{li-etal-2022-curriculum} or context-manipulation \cite{nagatsuka-etal-2021-pre,mohankumar-etal-2024-improving}. In-sample evaluation is prevalent in this quadrant: they raise the possibility that observed gains stem from the artificial variants themselves rather than their temporal ordering.
The rationale is consistently borrowed from human cognition: models benefit from progressing easy-to-hard \cite{tsvetkov-etal-2016-learning}, irregular patterns induce noisier gradients \cite{platanios-etal-2019-competence}, and scaffolding stabilises early optimisation \cite{nagatsuka-etal-2021-pre}. Notably, task-agnostic proxies are not task-neutral: their choice is typically guided by task-specific intuition.

\medskip
\noindent \textbf{TA-M} proxies infer difficulty from model-derived signals without requiring task labels. 
Three families appear: auxiliary models providing generic scores not directly related to the target task \cite{wang-etal-2019-dynamically,zhou-etal-2020-uncertainty}; leveraging pre-trained embedding geometry \cite{lu-lam-2023-pcc,weinzierl-harabagiu-2024-discovering,pan-etal-2025-negative}; and LLMs used directly as raters or to generate contrastive negatives of varying difficulty \cite{li-etal-2024-active,xu-etal-2024-automatic}. 
The shared motivation is to replace human assumptions with reusable model-based signals. Less common than TA-H, this quadrant has grown more accessible with the rise of strong pre-trained encoders.

\medskip
\noindent \textbf{TD-H} proxies are tied to the target task but remain human-centred by either utilising human-created references or directly leveraging human performance on the target task. A first family defines difficulty directly in target space \cite{kocmi-bojar-2017-curriculum,kano-etal-2021-quantifying,wang-etal-2024-theoremllama}. A second family injects target information into the input to create a difficulty scaffolding during training \cite{stojanovski-fraser-2019-improving,wei-etal-2021-trigger}. A third defines difficulty by considering the discrepancy to human references \cite{magooda-litman-2021-mitigating-data,agrawal-carpuat-2022-imitation}. Finally, a fourth directly captures human performance, preference, or disagreement on the task \cite{elgaar-amiri-2023-hucurl,hamad-etal-2024-fire,pattnaik-etal-2024-enhancing}. TD-H proxies draw on parallels to human learning, expect target-related guidance to ease optimisation early in training, or aim to leverage inherent task structure for smoother training.

\medskip
\noindent \textbf{TD-M} is the most common quadrant and most adaptive to the model: difficulty is defined through model-data interaction, evaluated globally or locally during training. Approaches use the current model to serve as evaluator \cite{sachan-xing-2016-easy,feng-etal-2019-learning,wan-etal-2020-self}; auxiliary evaluators to estimate locally \cite{liu-etal-2021-scheduled,yu-etal-2024-popalm} or globally \cite{su-etal-2021-dialogue,pattnaik-etal-2024-enhancing}; or teacher models to derive difficulty scores \cite{shen-feng-2020-cdl,lobov-etal-2022-applying,christopoulou-etal-2022-training}. 
A repeated motivation is dissatisfaction with rigid human heuristics and the desire for fully automated pacing. The tight coupling between difficulty estimation and scheduler dynamics is the defining feature of TD-M and its main interpretive challenge.

\medskip
\noindent \textbf{Mixtures of quadrants} appear in few works, either combining proxies sequentially \cite{liu-etal-2023-code,ranaldi-etal-2024-language} or aggregating multiple signals \cite{dou-etal-2020-dynamic}. TA-H measures most often serve as the inexpensive backbone, enriched by more adaptive signals. Thus, quadrants constitute composable sources of difficulty information, not mutually exclusive strategy types.

Across quadrants, some proxies target data usefulness rather than inherent task difficulty, treating noisy, unreliable, or out-of-domain instances as hard \cite{luo-etal-2017-learning-noise,wang-etal-2019-dynamically,shamanna-girishekar-etal-2021-training}; such cases blur difficulty progression with denoising, filtering, or domain selection.

\subsection{Survey of Training Schedulers}
\label{sec:survey-scheduler}
We now survey how schedulers translate difficulty evaluations into the expected training contribution of a sample, organised by retention regime (\Cref{sec:scheduler-retention}) and using selection, frontier, and exposure monotonicity as the main comparative lens. We focus on the main approaches in each retention regime; see \Cref{sec:extended-scheduler-survey} for an extended account of the literature and examples.

\begin{figure}
    \includegraphics[width=\linewidth]{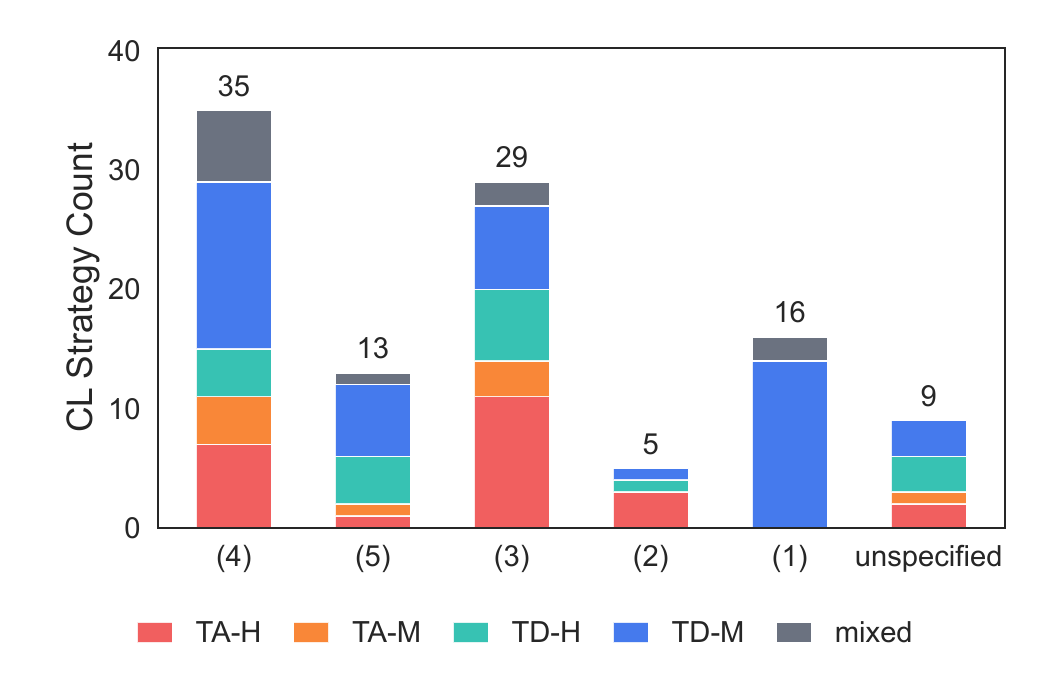}
    \caption{Distribution of quadrants per retention regime. The numbering follows the main paper: (1) unconstrained, (2) no re-entrance, (3) no retention, (4) full retention, (5) universal.}
        \label{fig:survey-eval}
\end{figure}

\medskip
\noindent \textbf{\Unconstrained{}} schedulers impose no structural restriction on how $A_t$ evolves. 
This mainly arises under local difficulty evaluation, where instances are repeatedly re-ranked under the current model state and $A_t$ reconstructed stage by stage \cite{pattnaik-etal-2024-enhancing,feng-etal-2019-learning,ruiter-etal-2020-self}; it also occurs when thresholds for accepting an instance into $A_t$ depend on current model performance \citep{xu-etal-2020-dynamic,lalor-yu-2020-dynamic} or policies select among fixed buckets without continuity constraints \cite{kumar-etal-2019-reinforcement}.
Such schedulers are often adaptive, but structurally guarantee neither selection, frontier, nor exposure monotonicity.

\medskip
\noindent \textbf{\NoReentry{}} allows contiguous retention but forbids re-admission once an instance leaves $A_t$. 
It is rare and mainly realised through targeted removal, such as training on a high resource language before continuing with the target language \citep{elhady-etal-2025-emergent}, progressively dropping low scoring instances (implementing anti-curriculum) \citep{kano-etal-2021-quantifying}, or retaining a fixed subset while replacing the remainder \citep{ren-etal-2025-step}. Observed examples are monotone in selection, frontier, and exposure, but the regime itself does not guarantee this; monotonicity depends on which instances are kept or removed and whether they follow an ordered notion of difficulty or quality \cite{wu-etal-2024-instruction,liu-etal-2023-code}.

\medskip
\noindent \textbf{\NoRetention{}} requires each instance to contribute at most once across stages. This is typically realised by traversing disjoint difficulty buckets or strictly ordered one-pass datasets \cite{Cirik2016,wei-etal-2021-shot}, most often with global \cite{li-etal-2022-curriculum,ranaldi-etal-2023-modeling,ren-etal-2025-step} or in-sample evaluation \cite{nagatsuka-etal-2021-pre,wang-etal-2025-knowledge}. This usually leads to monotonicity in selection, frontier, and exposure, but is not guaranteed by the regime: reordering within future buckets \cite{zhang-etal-2025-learning-like} or epoch-wise curriculum resets yields overall non-monotone schedules \cite{kocmi-bojar-2017-curriculum,kumari-etal-2021-sentiment}.

\medskip
\noindent \textbf{\FullRetention{}} keeps instances available once admitted. This is the dominant retention regime, often motivated by smoother optimisation and reduced forgetting relative to disjoint stage-wise training.
Canonical examples are competence-based \cite{platanios-etal-2019-competence} and Baby-Step schedulers \cite{spitkovsky-etal-2010-baby}, which expand $A_t$ cumulatively over an ordered ranking and are thus structurally frontier monotone; under aligned selection and uniform sampling, exposure monotonicity follows \cite{chang-etal-2021-order,wu-etal-2023-empower,yao-etal-2025-language-models}.
Variants differ mainly in pacing \cite{wang-etal-2022-feeding,zhou-etal-2020-uncertainty}, batching \cite{liu-etal-2024-fisher}, or defining difficulty thresholds \citep{yadav-etal-2017-learning,yuan-etal-2022-generative-entity}. Combining sampling and weighting need not alter these structural properties when both mechanisms are aligned \cite{liu-etal-2020-norm}.
Since \fullretention{} constrains only support evolution, not the direction or distribution of contribution mass, it preserves frontier monotonicity under local evaluation but not selection or exposure monotonicity \cite{tang-etal-2025-effective,sachan-xing-2016-easy}; ordered training within $A_t$ \citep{li-etal-2024-active} or epoch-wise resets \cite{lee-etal-2022-efficient-pre} can likewise render training non-monotone.

\medskip
\noindent \textbf{\UniversalA{}} schedulers keep all instances admissible throughout training and progression is realised solely by redistributing contribution mass, rendering the notion of difficulty frontier uninformative. 
This regime appears both in weighting-based curricula that shift mass towards harder instances \cite{yang-etal-2022-take} and in sampling-based curricula with time-varying batch mixtures over dataset partitions \cite{du-etal-2019-extracting,zhang-etal-2025-preference}. Irrespective of the employed mechanism, these examples are monotone in selection and exposure, but the regime itself does not guarantee it:
redistribution of gradient mass within the mini-batch \cite{wan-etal-2020-self, liu-etal-2024-curriculum} yields no structural monotonicity.

\section{Discussion}
\label{sec:discussion}
Our careful analysis shows that the difficulty of interpreting CL results in NLP is not merely due to empirical variation, but to three recurring conflations: what counts as difficult, how progression is scheduled, and which objective the curriculum is meant to serve.
These conflations impede meaningful comparison across approaches. In the following, we discuss how the taxonomy makes these tensions visible and why they matter methodologically.

\medskip
\noindent \textbf{Difficulty Evaluation.}
The survey shows that difficulty functions in NLP are highly heterogeneous. Our taxonomy structures this design space via axes not only capturing technical properties, but also assumptions about what renders an instance difficult and why ordering by that signal should improve training.
Readability scores, linguistic similarity, perplexity, and loss each locate difficulty differently: 
in the data, the model, or their interaction. Empirical evidence further suggests that different difficulty proxies disagree on which instances are difficult \cite{christopoulou-etal-2022-training,Toborek2026}. Positive results across proxies can thus coexist without any of them confirming the same underlying mechanism. In-sample difficulty evaluation further sharpens this concern: defining difficulty by a function transforming the instances may result in
ordering that does not recover an inherent task progression, but instead
introduces instances deviating from the original data distribution. In this sense, in-sample evaluation is similar to data augmentation with temporally controlled data exposure.

As a consequence, future work needs to state its rationale for proxy choice to allow for better interpretation and comparison of results. In-sample evaluations need to compare against a baseline that contains the transformed instances, rather than comparing only against the original dataset.

\medskip
\noindent \textbf{Scheduling.}
The survey supports the diagnosis that schedulers have been systematically underspecified, often independently of the difficulty function they accompany. The taxonomy addresses this by formalising scheduler progression via mechanism-independent components: retention regimes and monotonicity properties.
Schedulers described as monotone need not be monotone in selection, frontier, and exposure simultaneously; thus, successful schedulers may in fact realise different forms of progression. Moreover, easy-to-hard schedulers can be implemented under different retention regimes. Both a \fullretention{} and a \noretention{} regime may be exposure monotone, while exposing the model to substantially different training distributions. While some work compares schedulers directly, these comparisons typically vary the direction of progression \cite{elgaar-amiri-2023-ling} or different pacing for the same scheduler \cite{platanios-etal-2019-competence}. Retention therefore remains a central but neglected design variable, despite distribution shifts being a core concern in continual learning \cite{Wang2024continualsurvey}.
This underspecification compounds under local difficulty evaluation: apparent monotonicity may  stem from the evolution of the difficulty function, the scheduler logic, or both. 

Consequently, schedulers should be described precisely in terms of retention regime and monotonicity type. 
Ablations should vary these dimensions independently of the difficulty proxy; under local difficulty functions, the evolution of $d_t$ should be reported separately from the scheduler.

\medskip
\noindent \textbf{Objectives.}
Applying the taxonomy surfaces a third conflation: many works labelled as CL pursue goals like data cleaning or domain filtering, which do not aim to leverage an inherent task progression \citep{Bengio2009}, but assume that some instances are less useful for training. Intuitively, noisy or out-of-domain data may count as difficult, yet one can equally imagine datasets which are noisy, contain out-of-domain instances \textit{and} would benefit from an ordering that targets some inherent task scaffolding. 
A related conflation concerns whether CL is used throughout training or only as a warm-start: these settings make different claims about the usefulness of the curriculum.

As a consequence, future work should analyse whether a  strategy succeeds because of its objective, the dataset, or their interaction, and compare strategies with distinct objectives.

\subsection{Future Directions}
Progress in CL will require moving from method-level comparison to component-level evaluation to better understand difficulty proxies, schedulers, and training objective interplay. Broad empirical benchmarks are valuable, but they cannot localise where observed gains stem from \cite{Zhou2024CurBench}.
Similarly, experiments that vary these components independently would make both positive and negative findings \cite{surkov2022do,weber2023curriculum} easier to interpret.

Our analysis further highlights underexplored design choices, especially the \universala{} and \noreentry{} regimes. Likewise, TA-M proxies and LLM-based judges offer scalable, label-free difficulty estimates, but should be evaluated by the assumption they encode rather than as generic stronger heuristics. The proposed taxonomy provides the framework for designing such targeted studies.

\section{Conclusion}
\label{sec:conclusion}

We identified an incomparability problem in the results of CL in NLP: works under the same CL label often differ in how they define difficulty, realise training progression, and motivate the curriculum objective. We addressed this with a two-part taxonomy that characterises difficulty evaluation by attribution source and task dependence, and unifies different scheduler implementations under their expected training contribution. Applied to \numpapers{} papers, the taxonomy makes structural differences between CL strategies explicit and enables improved reporting, ablation, and interpretation of future results.

\section{Limitations}
Although the proposed taxonomy is in principle applicable to any data-level CL strategy, we have only applied it to the corpus based on the ACL anthology. We are confident the taxonomy will be applicable to the broader CL literature, but this remains to be tested. The taxonomy is analytical in exposing structural differences between difficulty functions and schedulers, but it does not predict which strategy will be effective for a given model, dataset, or objective. \Cref{sec:discussion} offers concrete recommendations on how future work should analyse the efficacy of CL design choices. 

While the taxonomy cleanly separates difficulty evaluation from scheduling, it does not yet provide an equally accessible way to assess the objective of a CL strategy. Our survey identifies objective-level conflations, such as difficulty progression versus filtering or warm-starting, but a systematic exploration of CL objectives remains future work.
Finally, several classifications required interpretive judgement, since many papers do not fully specify their difficulty rationales or scheduling mechanisms. We therefore classify monotonicity conservatively, only when it is structurally guaranteed, which may undercount approaches that are empirically monotone in specific implementations.

\section*{Acknowledgments}
We would like to thank Katharina Beckh for her support and the insightful discussions.

\bibliography{anthology-shrunk,custom}

\appendix
\clearpage

\section{Corpus Compilation and Annotation}
\label{sec:corpus-compilation}
We use the public ACL Anthology API to compile the paper corpus. Initially, we queried titles and abstracts for either the exact phrase \emph{``curriculum learning''} or the co-occurrence of \emph{``curriculum''} (or \emph{``curricula''}) and \emph{``learning''}. This broad query intentionally prioritised recall, yielding a heterogeneous set of candidate papers spanning multiple modalities and application domains.

In a first automated filtering pass, we restricted the corpus to papers published at venues classified as \emph{conferences} in the ACL Anthology, which includes works published in Findings.
We then applied keyword-based filters to only retain text-only tasks removing work primarily concerned with medical imaging, multimodal learning, robotics, and speech processing. This resulted in a set of $199$ papers. In a second, manual review stage, we excluded work that was not written in English, focused on educational curricula rather than curriculum learning for neural network training, or applied CL at levels other than the data-level (cf. \Cref{sec:background}). 
Candidate papers were distributed across the authors for manual assessment.
Cases where the inclusion or annotation were uncertain, were discussed jointly until consensus was reached.
Because disagreements were resolved during annotation rather than recorded through an independent double-annotation procedure, we do not report inter-annotator disagreement.
The remaining papers constitute the final dataset analysed in this survey.

For the annotation of scheduler properties, a \checkmark{} was assigned only when the scheduler guaranteed the corresponding monotonicity by construction for every pair of consecutive stages.
We inspected the method description, equations, pseudocode, and implementation details where available.
When the property depended on unspecified implementation details, stochastic outcomes, or the evolution of a local difficulty function, we did not assign a \checkmark{}.

The remaining papers constitute the final dataset analysed in this survey.
While the final set comprises \numpapers{} papers, several papers propose multiple distinct strategies, yielding a total of \numstrategies{} annotated strategies.

\section{Computing and Interpreting $\pi_t$}
\label{sec:example-scheduler}

To illustrate how the scheduler taxonomy can be applied in practice, consider the SST-2 \citep{Socher2013} training set with $n=67\,349$ training instances, ordered by increasing $d_t$ defined as the tokenised input length.
The \piname{} was deliberately defined abstractly, and its derived properties allow direct comparison of schedulers without knowledge of their concrete implementations.
However, if we want to compute $\pi_t$ explicitly, we need to specify how the scheduler is implemented.
The pacing function determines how quickly the curriculum advances through the difficulty-ranked dataset by mapping each stage to the highest rank reached at that stage.
In this example, suppose the curriculum is defined over $T=5$ stages with equally sized increase sets $C_t$ yielding $C_t = {x_i: c_{t-1} \leq i \leq c_t}$, such that

\[
\begin{aligned}
(c_1,\ldots,c_5) = (&13\,470,\;26\,940,\;40\,410,\\
                    &53\,880,\;67\,349).
\end{aligned}
\]

For simplicity, we consider a pure sampling scheduler, where all admissible instances receive equal loss weight, i.e. $w_t(x)=1$.
Consequently,
$\pi_t(x)=p_t(x)$.

Now compare two schedulers that use the \emph{same} difficulty function and pacing schedule but differ only in their retention regime.

\begin{figure*}
   \centering
   \includegraphics[width=\textwidth]{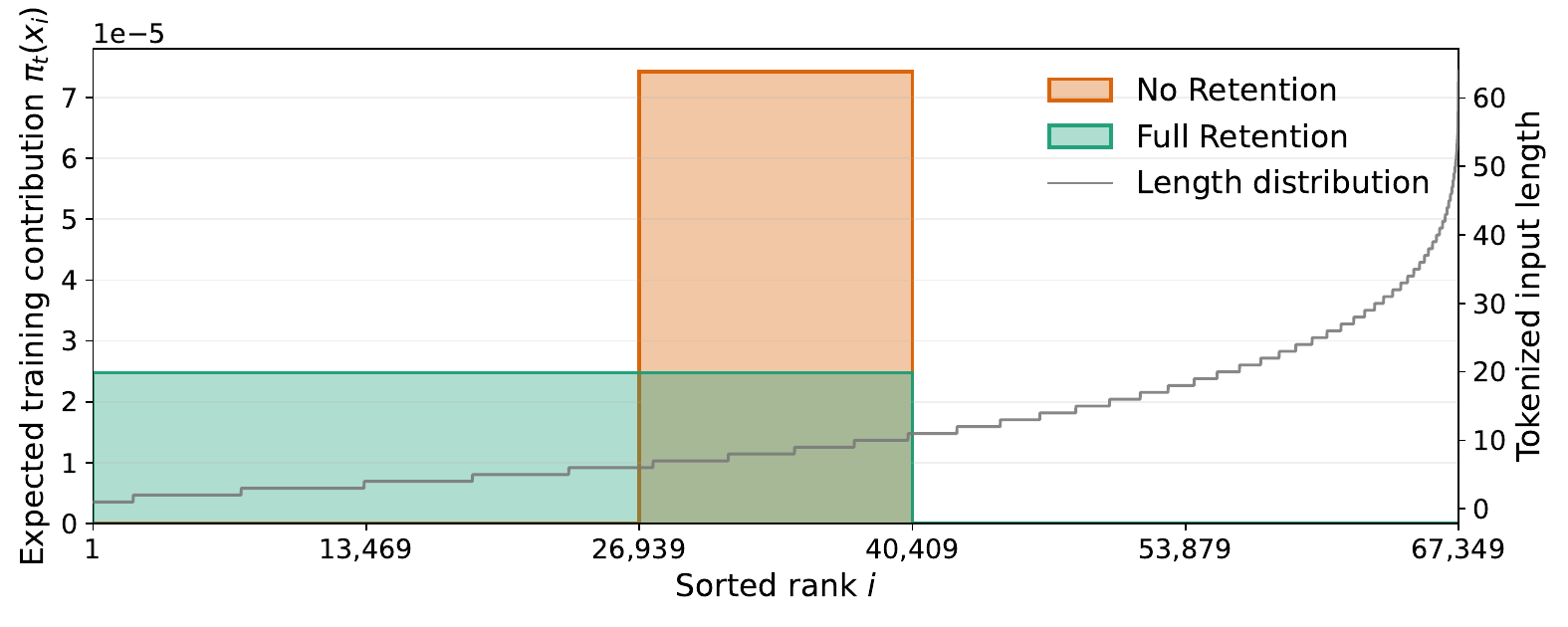}
   \caption{Expected training contribution $\pi_3(x)$ at stage $t=3$ for no- and full-retention schedulers with the same length-based difficulty ordering and pacing; grey line visualises tokenised input length. Both reach the same curriculum frontier, but no retention concentrates probability mass on newly admitted instances while full retention spreads it across all admitted instances. Identical ordering and pacing may still produce different training distributions.}
   \label{fig:c3-scheduler-comparison}
\end{figure*}

\paragraph{No-Retention}
A no-retention scheduler, such as that of \citet{Bengio2009}, exposes only the newly admitted instances at each stage:
\[
A_t^{\mathrm{NR}}
=
C_t.
\]
Sampling is uniform over the admissible set, giving

\[
\pi_t^{\mathrm{NR}}(x)
=
\begin{cases}
\dfrac{1}{|A_t^{\mathrm{NR}}|},
&
x \in A_t^{\mathrm{NR}},
\\[6pt]
0,
&
\text{otherwise}.
\end{cases}
\]
At stage $t=3$, only instances ranked $26\,941$ through $40\,410$ are sampled.
Each admissible instance therefore receives probability
\[
\pi_3^{\mathrm{NR}}(x)
=
\frac{1}{13\,470}
\approx
7.42\times10^{-5}.
\]

\paragraph{Full-Retention}
A full-retention scheduler, such as the competence-based scheduler of \citet{platanios-etal-2019-competence}, instead keeps all previously introduced instances available:
\[
A_t^{\mathrm{FR}}
=
\bigcup^t_{i=1}C_i,
\]
leading to
\[
\pi_t^{\mathrm{FR}}(x)
=
\begin{cases}
\dfrac{1}{|A_t^{\mathrm{FR}}|},
&
x \in A_t^{\mathrm{FR}},
\\[6pt]
0,
&
\text{otherwise}.
\end{cases}
\]

At stage $t=3$, the scheduler samples uniformly from the first $40\,410$ instances, so that
\[
\pi_3^{\mathrm{FR}}(x)
=
\frac{1}{40\,410}
\approx
2.47\times10^{-5}.
\]

Although both schedulers use the same difficulty function and pacing schedule and reach the same curriculum frontier at every stage, they induce different training distributions.
\Cref{fig:c3-scheduler-comparison} visualises the expected training contribution $\pi_3$ under both schedulers for stage $t=3$.
Under no-retention, all probability mass is concentrated on the newly introduced portion of the curriculum, whereas under full-retention it is distributed across all previously admitted instances.
This example illustrates how the proposed formalism makes scheduler behaviour explicit through $\pi_t$, revealing implementation differences that remain hidden under the common description of both strategies as ``easy-to-hard'' curricula.


\subsection{Relation to Practical Implementations}
Our proposed formalism characterises the curriculum scheduler independently of the optimisation algorithm. 
Optimiser state and learning-rate schedules affect the magnitude of parameter updates induced by sampled instances, but not the scheduler itself. Extending the framework to explicitly model these optimisation dynamics would therefore constitute a natural step towards a predictive theory of curriculum learning.

Batch construction, by contrast, is already captured through the sampling distribution $p_t$. 
Any batching strategy induces a marginal probability that an instance $x$ is sampled at stage $t$, and this probability can be incorporated directly into $\pi_t$. 
Likewise, implementations using stages of unequal duration simply accumulate different numbers of samples from the corresponding stage-specific distributions. 
While this changes the cumulative exposure of individual training instances, it does not affect the scheduler properties defined with respect to $\pi_t$, including the proposed retention regimes and monotonicity properties.

\section{Extended Survey}
\label{sec:extended-survey}

This appendix provides an extended account of the strategies identified in our survey, complementing the aggregate results presented in the main paper. We first examine the difficulty proxies used across the four quadrants and then turn to the scheduler structures through which they control training exposure.

\subsection{Difficulty Quadrants}
\label{sec:extended-quad-survey}

We first examine how difficulty is operationalised across the four quadrants, focusing on the proxies used in the literature and the motivations given for their use.

\begin{figure}
    \centering
    \includegraphics[width=\linewidth]{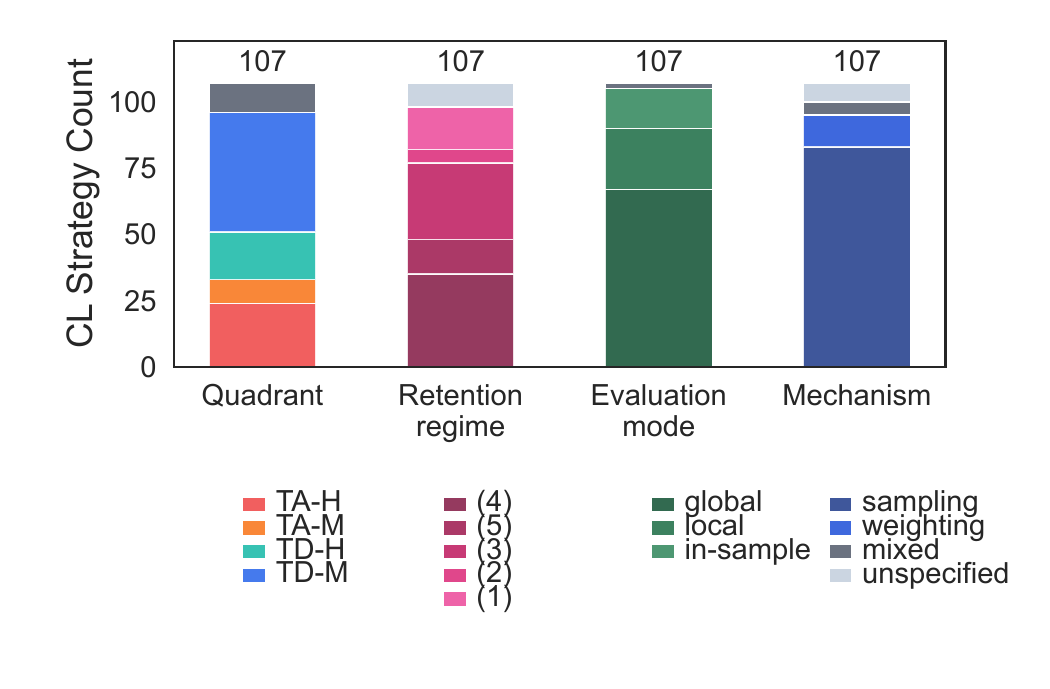}
    \caption{Overview of counts for quadrants, retention regime, evaluation mode, and scheduler mechanism.}
    \label{fig:counts}
\end{figure}

\begin{figure}
    \centering
    \includegraphics[width=\linewidth]{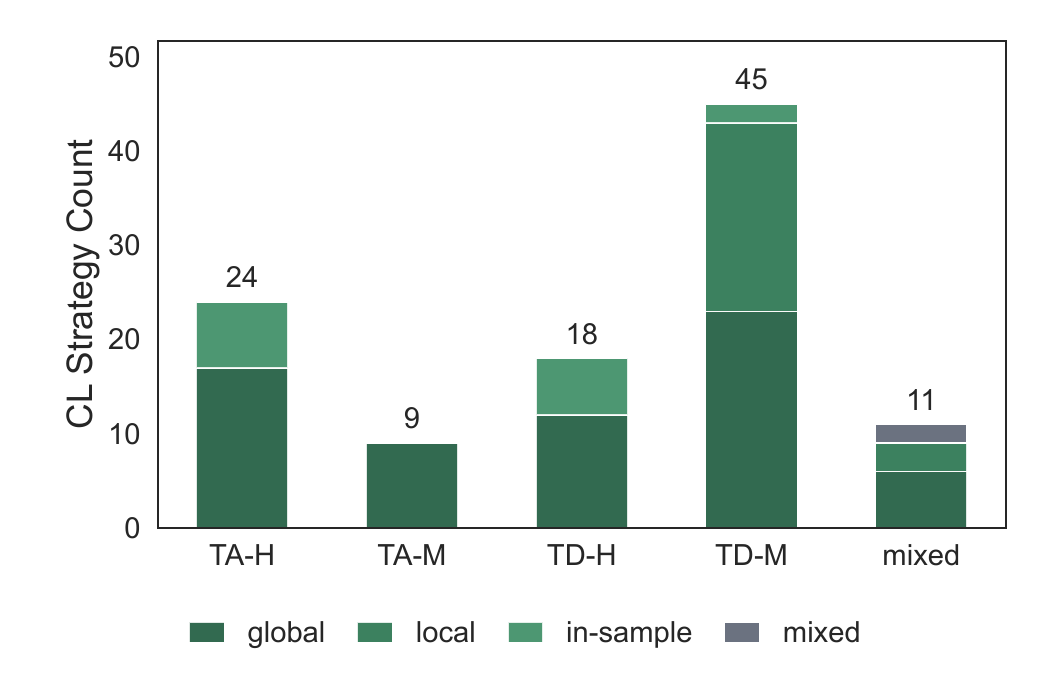}
    \caption{Counts for evaluation mode per quadrant.}
    \label{fig:mode-per-quadrant}
\end{figure}

\Cref{fig:counts,fig:mode-per-quadrant} provide aggregate counts and show how the quadrants relate to evaluation mode.

\subsubsection{TA-H (Task-agnostic Human-Centric) }
\label{sec:ta-h}

TA-H is the most accessible quadrant: proxies are human-designed, label-free properties of the input, making them inexpensive to compute. They are often treated as default curricula. At the same time, they are frequently chosen through task-specific human intuition, so ``task-agnostic'' may still not be task-neutral in practice.

Most TA-H curricula rely on psycholinguistic and readability-related proxies such as input length \cite{platanios-etal-2019-competence,lobov-etal-2022-applying} and syntactic or shallow structural complexity \cite{tsvetkov-etal-2016-learning,chen-etal-2022-crossroads}. A second family uses source-level attributes, including genre \cite{elzohbi-zhao-2025-tahdib}, dataset tier \cite{wu-etal-2024-instruction,hong-etal-2025-game}, or language-level factors such as availability \cite{elbayad-etal-2023-fixing} and linguistic similarity \cite{shamanna-girishekar-etal-2021-training} in multilingual transfer. A third line controls difficulty by construction, e.g. through explicit compositional depth \cite{wu-etal-2023-empower,yao-etal-2025-language-models}, constrained reasoning structure \cite{ren-etal-2025-step}, or targeted perturbations \cite{li-etal-2022-curriculum,kakkar-etal-2023-search} and context manipulation \cite{nagatsuka-etal-2021-pre,mohankumar-etal-2024-improving}.

The dominant reason for TA-H is that models, like human learners, may benefit from progressing from simpler and more prototypical inputs to more complex ones \cite{tsvetkov-etal-2016-learning,lee-etal-2022-efficient-pre,toborek-etal-2025-beyond}.
Other work adopts a statistical perspective: infrequent or structurally irregular patterns induce noisier gradients due to limited evidence \cite{platanios-etal-2019-competence,yadav-etal-2017-learning}. 
A third recurring argument appeals to information-processing limits: longer contexts, rarer phenomena, and deeper compositional chains are assumed to impose higher computational load and to destabilise optimisation early in training \cite{platanios-etal-2019-competence,li-etal-2022-complex,yao-etal-2025-language-models}. 
Finally, some curricula use human-designed scaffolding to improve training of neural architectures, temporarily simplifying the input \cite{nagatsuka-etal-2021-pre,wang-etal-2025-knowledge} or enriching context \cite{mohankumar-etal-2024-improving} before gradually removing these aids to encourage generalisation.

\subsubsection{TA-M (Task-agnostic Model-Centric)}
\label{sec:ta-m}

TA-M proxies infer difficulty from model-derived signals independent of the target task. They are less common than TA-H, but the rise of pre-trained embeddings and foundation models has made this quadrant increasingly accessible. It replaces handcrafted assumptions with model-based estimates while remaining independent of task labels.

We identify three main families of TA-M proxies. First, some approaches  use auxiliary models not trained on the target task to provide generic difficulty cues such as surprisal \cite{lobov-etal-2022-applying,zhou-etal-2020-uncertainty} or token-level cross-entropy \cite{wang-etal-2019-dynamically}. Second, several works derive difficulty from pre-trained representation spaces, e.g. through static word-vector norms \cite{liu-etal-2020-norm} or distances and similarities between model embeddings, including Euclidean distance \cite{weinzierl-harabagiu-2024-discovering}, cosine similarity \cite{pan-etal-2025-negative}, and BLEURT-based similarity \cite{lu-lam-2023-pcc}. Third, recent work delegates difficulty estimation to LLMs, either by directly rating the difficulty of an input \cite{li-etal-2024-active} or by using strong and weak LLMs to generate contrastive negatives of varying difficulty for post-training \cite{xu-etal-2024-automatic}.

A central reason for TA-M is that difficulty should be defined from the perspective of the model; model-centric signals are therefore used as reusable estimates of processing difficulty \cite{zhou-etal-2020-uncertainty,li-etal-2024-active}. In contrastive learning and data augmentation, where semantic distance or compatibility between instances directly shapes training difficulty, embedding-based similarity is a natural control variable for constructing informative rather than misleading pairs \cite{lu-lam-2023-pcc,weinzierl-harabagiu-2024-discovering,pan-etal-2025-negative}. Finally, several works aim to stabilise training by delaying training with uncertain, weakly aligned, or semantically incompatible examples \cite{wang-etal-2019-dynamically,lobov-etal-2022-applying}.

\subsubsection{TD-H (Task-dependent Human-Centric)}
\label{sec:td-h}
TD-H proxies are tied to the target task, human derived, and often interpretable signals. They often exploit interpretable task structure, but the quadrant includes both genuine human difficulty signals and explicit scaffolding strategies simplifying the task early in training.

Some works use target complexity directly as proxy, e.g. through target length \citep{kocmi-bojar-2017-curriculum,kano-etal-2021-quantifying,jia-etal-2023-sample}, code complexity in terms of required conditional statements \citep{dou-etal-2024-stepcoder}, AMR graph depth \citep{wang-etal-2022-hierarchical}, or number of proof steps \citep{wang-etal-2024-theoremllama}. 
Others define difficulty by injecting target information to the input, such as oracle pronouns \citep{stojanovski-fraser-2019-improving}, gold-arguments \citep{wei-etal-2021-trigger}, or prompt-level target injection \citep{yang-song-2022-fpc}. 
A third set uses human-reference-based discrepancy signals, including edit-distance measures \citep{chang-etal-2021-order,agrawal-carpuat-2022-imitation}, ROUGE score \cite{magooda-litman-2021-mitigating-data}, retrieval-based answerability estimates \citep{tay-etal-2019-simple}, or target-centred graph structure \citep{vakil-amiri-2023-curriculum}.
Finally, some works directly capture human performance proxies like their annotation time \cite{hamad-etal-2024-fire} or human preference \citep{pattnaik-etal-2024-enhancing}.

Three reasons for TD-H recur.
First, some approaches draw on parallels to human learning, either through psycholinguistic development \citep{singh-etal-2018-curriculum,wang-etal-2022-hierarchical,wang-etal-2024-theoremllama} or direct proxies of human processing cost \cite{hamad-etal-2024-fire}. 
Second, many works use TD-H proxies to ease optimisation early in training by providing target-related guidance before gradually transitioning to the full task \cite{stojanovski-fraser-2019-improving,yang-song-2022-fpc,dou-etal-2024-stepcoder}. 
Third, other works assume that an inherent structure in the learning problem can be leveraged for smoother training \citep{agrawal-carpuat-2022-imitation,grishina-sorokin-2022-local,pattnaik-etal-2024-enhancing}.

\subsubsection{TD-M (Task-dependent Model-Centric)}
\label{sec:td-m}
TD-M defines difficulty through model-data interaction, most often in a local and optimisation-coupled manner. This makes it the most adaptive quadrant, but also the one where difficulty evaluation and scheduler dynamics are most tightly entangled. Across both local and global approaches, this is the most common type of difficulty proxy.

One group uses the \textit{current model} itself as evaluator, deriving difficulty proxies from representation similarity \citep{ruiter-etal-2020-self}, confidence or uncertainty \citep{wan-etal-2020-self,tang-etal-2025-effective}, and criteria based on loss, margin, or error \citep{feng-etal-2019-learning,huang-du-2019-self,sachan-xing-2016-easy}. One global variant estimates difficulty from the pre-trained model using Fisher-information-based sensitivity scores \citep{liu-etal-2024-fisher}.
A second group relies on \textit{auxiliary evaluation models}. Locally, they estimate sample complexity \citep{yu-etal-2024-popalm} or learning progress during training \citep{liu-etal-2021-scheduled,xu-etal-2020-dynamic}; globally, they provide fixed task-specific scores such as preference rankings \citep{pattnaik-etal-2024-enhancing}, relevance estimates \citep{su-etal-2021-dialogue}, or appropriateness judgments \citep{kano-etal-2021-quantifying}.
A third group derives global difficulty scores using signals from \textit{teacher models} such as prediction confidence or probability \citep{shen-feng-2020-cdl,lobov-etal-2022-applying} or training-dynamics summaries \citep{poesina-etal-2024-novel,christopoulou-etal-2022-training}, or local difficulty scores using directly task-related measures like question answering probability \citep{maharana-bansal-2022-curriculum}.
Finally, some approaches make difficulty \textit{model-conditioned at the sample level}, e.g. by perturbing the embedding space \citep{wang-etal-2025-knowledge} or generating examples closer to the decision boundary \citep{jin-etal-2024-enhancing}.

The reasons why several works turn to TD-M is criticism of the inaccuracy \cite{wan-etal-2020-self,christopoulou-etal-2022-training} and rigidity \cite{liu-etal-2024-curriculum} of human heuristics. Others aim to control data reliability more finely, e.g. under noisy supervision \cite{shen-feng-2020-cdl,kano-etal-2021-quantifying} or the presence of false negatives \cite{su-etal-2021-dialogue,feng-etal-2019-learning}. A third reason is automation: using model-based signals for both difficulty estimation and pacing rather than hand-designing curricula \cite{kumar-etal-2019-reinforcement,wang-etal-2019-dynamically}. Finally, others use TD-M proxies to improve optimisation by tailoring training exposure to the current model \cite{maharana-bansal-2022-curriculum,jin-etal-2024-enhancing}.

\subsubsection{Mixture of Quadrants}
\label{sec:quadrant-mix}
Not all curriculum strategies rely on a single quadrant. A small number of works combine proxies from different quadrants, either sequentially, by using one signal for coarse staging and another for finer ordering \cite{liu-etal-2023-code,ranaldi-etal-2024-language}, or jointly, by aggregating multiple signals into one difficulty score \cite{luo-etal-2017-learning-noise,dou-etal-2020-dynamic,dai-etal-2021-preview}. Most often, TA-H measures serve as inexpensive and widely available building blocks that are enriched by more task- or model-specific information.

These mixtures suggest that the quadrants are best understood as composable sources of difficulty information rather than mutually exclusive strategy types. This is especially useful when simple human-designed heuristics are available at scale but benefit from being complemented by more adaptive or task-specific signals.

\subsection{Training Schedulers}
\label{sec:extended-scheduler-survey}
We next examine how difficulty evaluations are translated into control over sample exposure during training.
We organise the surveyed strategies by the retention regimes defined in \Cref{sec:scheduler-retention}, using selection, frontier, and exposure monotonicity as the main comparative lens.

\medskip
\noindent \textbf{\Unconstrained{}} schedulers in the \unconstrained{} regime impose no structural restriction on how $A_t$ evolves. 
In NLP, this regime arises mainly under local difficulty evaluation, where instances are repeatedly re-ranked under the current model state and $A_t$ reconstructed stage by stage \cite{pattnaik-etal-2024-enhancing,feng-etal-2019-learning,ruiter-etal-2020-self}. 
It also appears irrespective of the evaluation mode when thresholding depends on current model performance \citep{xu-etal-2020-dynamic,lalor-yu-2020-dynamic} or when policies choose among fixed buckets without continuity constraints \cite{kumar-etal-2019-reinforcement}.
Such schedulers are often adaptive, but structurally guarantee neither selection, frontier, nor exposure monotonicity.

\medskip
\noindent \textbf{\NoReentry{}} allows contiguous retention but forbids re-admission once an instance leaves $A_t$. 
In NLP, this is mainly realised through targeted removal: e.g. training on a high resource language before continuing with the target language \citep{elhady-etal-2025-emergent} or progressively dropping low scoring instances (implementing anti-curriculum) \citep{kano-etal-2021-quantifying}. Complementary, a strategy can retain a fixed subset across stages while replacing the remainder \citep{ren-etal-2025-step}. In both cases, the curriculum is monotone in selection, frontier, and exposure; yet this is not guaranteed by the regime and depends entirely on which instances are kept or removed and whether they follow an ordered notion of difficulty or quality \cite{wu-etal-2024-instruction,liu-etal-2023-code}.

Overall, the \noreentry{} regime is rare although in principle it could also include general sliding window approaches.

\medskip
\noindent \textbf{\NoRetention{}} requires each instance to contribute at most once across stages. In NLP, it is typically realised by traversing disjoint difficulty buckets or, in the limiting case, a strictly ordered one-pass dataset traversal \cite{Cirik2016,wei-etal-2021-shot}.
The regime is most common with global \cite{li-etal-2022-curriculum,ranaldi-etal-2023-modeling,ren-etal-2025-step} or in-sample evaluation mode \cite{nagatsuka-etal-2021-pre,wang-etal-2025-knowledge}, leading to monotonicity in selection, frontier, and exposure. 

Yet, the \noretention{} regime alone does \textit{not} guarantee globally monotone progression: reordering within future buckets \cite{zhang-etal-2025-learning-like} or resetting the curriculum each epoch yields overall non-monotone schedules \cite{kocmi-bojar-2017-curriculum,kumari-etal-2021-sentiment,lu-lam-2023-pcc}.

\medskip
\noindent \textbf{\FullRetention{}} requires instances to remain available for the rest of training once admitted. This is the dominant retention regime in NLP, often motivated by smoother optimisation and reduced forgetting relative to disjoint stage-wise training.

Canonical examples are competence-based \cite{platanios-etal-2019-competence} and Baby-Step schedulers \cite{spitkovsky-etal-2010-baby}, which implement cumulative expansion over an ordered ranking and are thus structurally frontier monotone; under aligned selection and uniform sampling from $A_t$ follows exposure monotonicity \cite{chang-etal-2021-order,wu-etal-2023-empower,yao-etal-2025-language-models}.
Many variants retain this same structure while differing only in pacing, e.g. by conditioning expansion on model proficiency \cite{wang-etal-2022-feeding,zhou-etal-2020-uncertainty}, operating on batches rather than instances \cite{liu-etal-2024-fisher}, or defining an explicit difficulty threshold for weighting \citep{yadav-etal-2017-learning} as well as sampling schedulers \cite{yuan-etal-2022-generative-entity,zheng-etal-2022-distilling}. Combining sampling and weighting need not alter these structural properties when both mechanisms are aligned \cite{liu-etal-2020-norm}.

Still, \fullretention{} alone does not guarantee a fully monotone curriculum since it only constrains support evolution. Under local difficulty evaluation, frontier monotonicity remains intact while inhibiting selection and exposure monotonicity \cite{tang-etal-2025-effective,sachan-xing-2016-easy}. Even under global evaluation, training on $A_t$ in a specific order can break exposure monotonicity \citep{li-etal-2024-active}. Likewise, this regime also accommodates anti-curriculum variants, where progression runs from hard to easy \cite{maharana-bansal-2022-curriculum} or from distant domains towards the target domain \cite{shamanna-girishekar-etal-2021-training}. As with the \noretention{} regime, some works also employ the \fullretention{} regime only for one epoch, rendering the overall training non-monotone \cite{lee-etal-2022-efficient-pre}.

\medskip
\noindent \textbf{\UniversalA{}} schedulers keep all instances admissible throughout training and progression is realised solely through redistribution of contribution mass. Accordingly, the notion of difficulty frontier is uninformative, and the relevant question is whether the redistribution induces monotone changes in selection or exposure.

In NLP, this regime appears both in weighting-based curricula that progressively shift mass towards harder instances \cite{yang-etal-2022-take} and in sampling-based curricula whose batch composition changes via a time-varying mixture of different dataset partitions \cite{du-etal-2019-extracting,xu-etal-2024-automatic,zhang-etal-2025-preference}. Irrespective of the employed mechanism, these examples are monotone in selection and exposure, but the regime itself does not guarantee it.
\citet{elgaar-amiri-2023-ling} compare different progressions over the same dataset, including anti-curriculum, curriculum, centre-outwards schedules and show that progression that is monotonically increasing performs best.

A separate line of work redistributes gradient mass within the current mini-batch \cite{wan-etal-2020-self, liu-etal-2024-curriculum}, yielding neither a meaningful frontier nor structural monotonicity.

\clearpage
\onecolumn
\section{Paper-level Analysis}
\label{sec:paper-analysis}
\begingroup
\fontsize{9pt}{10pt}\selectfont
\setlength{\tabcolsep}{2pt}
\renewcommand{\arraystretch}{1.25}
\begin{longtable}{@{}
p{0.24\textwidth}
@{\hspace{6pt}}
p{0.03\textwidth}
p{0.08\textwidth}
p{0.25\textwidth}
@{\hspace{6pt}}
>{\centering\arraybackslash}
p{0.03\textwidth}
p{0.02\textwidth}
p{0.02\textwidth}
p{0.02\textwidth}
@{\hspace{10pt}}
p{0.21\textwidth}@{}}
\caption{Full paper-level analysis. Duplicate entries indicate distinct approaches within one work; missing details are marked as --. Evaluation modes (M.) are abbreviated as local (l), global (g), and in-sample (i). Schedulers are described by retention regime (R.) and selection (S.), frontier (F.), and exposure (E.) monotonicity; retention regime numbering follows the main paper: (1) unconstrained, (2) no re-entrance, (3) no retention, (4) full retention, (5) universal.}
\label{tab:table-summary} \\
\noalign{\vskip 6pt}
\hline
\noalign{\vskip 2pt}
Paper & \multicolumn{3}{c}{Difficulty Evaluation} & \multicolumn{4}{c}{CL Scheduler} & Task \\[2pt]
\cline{2-8}
\noalign{\vskip 2pt}
 & M. & Quadrant & Proxy & R. & S. & F. & E. & \\[2pt]
\hline
\endfirsthead
\multicolumn{9}{@{}l}{\textit{Table \thetable\ continued.}}\\
\hline
\noalign{\vskip 2pt}
Paper & \multicolumn{3}{c}{Difficulty Evaluation} & \multicolumn{4}{c}{CL Scheduler} & Task \\[2pt]
\cline{2-8}
\noalign{\vskip 2pt}
 & M. & Quadrant & Proxy & R. & S. & F. & E. & \\[2pt]
\hline
\endhead
\hline
\endfoot
\hline
\endlastfoot
\cite{sachan-xing-2016-easy} & l & TD-M & impact on training loss; task error; margin to decision boundary & (4) & \xmark & \cmark & \xmark & classification; question answering \\
\midrule
\cite{tsvetkov-etal-2016-learning} & g & TA-H & diversity, simplicity, prototypicality measures & (3) & \xmark & \xmark & \xmark & task-specific word representation learning \\
\midrule
\cite{kocmi-bojar-2017-curriculum} & g & TD-M & sentence length & (3) & \xmark & \xmark & \xmark & seq2seq; machine translation \\
\midrule
\cite{luo-etal-2017-learning-noise} & g+l & TA-H+ TD-M & prior-based reliability + noise estimation & (4) & \cmark & \cmark & \cmark & classification; relation extraction \\
\midrule
\cite{yadav-etal-2017-learning} & g & TA-H & average word log frequency & (4) & \cmark & \cmark & \cmark & classification \\
\midrule
\cite{singh-etal-2018-curriculum} & g & TD-H & input and output length & -- & -- & -- & -- & generation; sentence generation \\
\midrule
\cite{du-etal-2019-extracting} & i & TD-H & use of reference spans & (5) & \cmark & \cmark & \cmark & information extraction and sequence labelling \\
\midrule
\cite{feng-etal-2019-learning} & l & TD-M & model loss & (1) & \xmark & \xmark & \xmark & classification; response selection \\
\midrule
\cite{feng-etal-2019-learning} & l & TD-M & peer instance separation gap & (1) & \xmark & \xmark & \xmark & classification; response selection \\
\midrule
\cite{feng-etal-2019-learning} & l & TD-M & model score on negative sample & (1) & \xmark & \xmark & \xmark & classification; response selection \\
\midrule
\cite{huang-du-2019-self} & l & TD-M & noise level estimate & (1) & \xmark & \xmark & \xmark & classification; relation extraction \\
\midrule
\cite{huang-du-2019-self} & l & TD-M & model loss & (1) & \xmark & \xmark & \xmark & classification; relation extraction \\
\midrule
\cite{kumar-etal-2019-reinforcement} & g & TD-M & noise level estimate & (1) & \xmark & \xmark & \xmark & seq2seq; machine translation \\
\midrule
\cite{platanios-etal-2019-competence} & g & TA-H & sentence length; word rarity & (4) & \cmark & \cmark & \cmark & seq2seq; machine translation \\
\midrule
\cite{stojanovski-fraser-2019-improving} & i & TD-H & presence of gold standard reference pronouns & (2) & \cmark & \cmark & \cmark & classification; anaphora resolution \\
\midrule
\cite{tay-etal-2019-simple} & i+g & TA-H+ TD-H & chunk length + answerability of text chunk & (1) & \xmark & \xmark & \xmark & classification; reading comprehension over long narratives \\
\midrule
\cite{wang-etal-2019-dynamically} & g & TA-M & domain relevance & (3) & \cmark & \cmark & \cmark & seq2seq; machine translation \\
\midrule
\cite{wang-etal-2019-dynamically} & g & TD-M & sample cleanliness & (3) & \cmark & \cmark & \cmark & seq2seq; machine translation \\
\midrule
\cite{zhang-etal-2019-curriculum} & g & TD-M & language model cross-entropy-difference & (4) & \cmark & \cmark & \cmark & seq2seq; machine translation \\
\midrule
\cite{dou-etal-2020-dynamic} & l & TA-H+ TA-M & TF-IDF similarity + round-trip BLEU & (1) & \xmark & \xmark & \xmark & seq2seq; machine translation \\
\midrule
\cite{lalor-yu-2020-dynamic} & g & TD-M & difficulty via Item Response Theory & (1) & \xmark & \xmark & \xmark & classification; NLU \\
\midrule
\cite{liu-etal-2020-norm} & g & TA-M & sum of word vector norms & (4) & \cmark & \cmark & \cmark & seq2seq; machine translation \\
\midrule
\cite{ruiter-etal-2020-self} & l & TD-M & similarity of model sentence representations & (1) & \xmark & \xmark & \xmark & seq2seq; machine translation \\
\midrule
\cite{shen-feng-2020-cdl} & g & TD-M & prediction probability of the ground truth label & (4) & \cmark & \cmark & \cmark & seq2seq; emotion-controllable response generation \\
\midrule
\cite{wan-etal-2020-self} & l & TD-M & model confidence & (5) & \xmark & \xmark & \xmark & seq2seq; machine translation \\
\midrule
\cite{xu-etal-2020-dynamic} & l & TD-M & loss decline & (1) & \xmark & \xmark & \xmark & seq2seq; low-resource machine translation \\
\midrule
\cite{zhou-etal-2020-global} & g & TD-M & agreement between distantly supervised labels and model predictions & -- & -- & -- & -- & seq2seq; machine translation \\
\midrule
\cite{zhou-etal-2020-uncertainty} & g & TA-M & data uncertainty (source): perplexity & (4) & \cmark & \cmark & \cmark & seq2seq; machine translation \\
\midrule
\cite{zhou-etal-2020-uncertainty} & g & TA-M+ TD-H & data uncertainty (joint): perplexity + length of target sentence & (4) & \cmark & \cmark & \cmark & seq2seq; machine translation \\
\midrule
\cite{zhou-etal-2020-uncertainty} & g & TD-H & data uncertainty (target): length of target sentence & (4) & \cmark & \cmark & \cmark & seq2seq; machine translation \\
\midrule
\cite{chang-etal-2021-order} & g & TD-H & distance between input and target & (4) & \cmark & \cmark & \cmark & generation; data-to-text \\
\midrule
\cite{dai-etal-2021-preview} & g & TA-H+ TD-M & rule-based complexity + auxiliary model performance & (4) & \cmark & \cmark & \cmark & classification; dialogue state tracking \\
\midrule
\cite{kano-etal-2021-quantifying} & g & TD-M & appropriateness & (2) & \cmark & \cmark & \cmark & seq2seq; summarisation \\
\midrule
\cite{kumari-etal-2021-sentiment} & g & TD-M+ TD-M & teacher model SBLEU + sentiment classifier score & (3) & \xmark & \xmark & \xmark & seq2seq; machine translation \\
\midrule
\cite{liu-etal-2021-scheduled} & l & TD-M & dialogue state complexity estimation & (1) & \xmark & \xmark & \xmark & generation; dialogue system \\
\midrule
\cite{magooda-litman-2021-mitigating-data} & g & TA-H & text specificity proxy & -- & \cmark & \cmark & \cmark & seq2seq; summarisation \\
\midrule
\cite{magooda-litman-2021-mitigating-data} & g & TD-H & ROUGE score & -- & \cmark & \cmark & \cmark & seq2seq; summarisation \\
\midrule
\cite{nagatsuka-etal-2021-pre} & i & TA-H & block-size of text input & (3) & \cmark & \cmark & \cmark & pre-training \\
\midrule
\cite{shamanna-girishekar-etal-2021-training} & g & TA-H & lexical similarity via provided vocabulary overlap & (4) & \cmark & \cmark & \cmark & classification \\
\midrule
\cite{shamanna-girishekar-etal-2021-training} & g & TD-M & classifier confidence & (4) & \cmark & \cmark & \cmark & classification \\
\midrule
\cite{su-etal-2021-dialogue} & g & TD-M & auxiliary model relevance score & (4) & \cmark & \cmark & \cmark & classification; response selection \\
\midrule
\cite{wang-etal-2021-progressive} & g & TD-M & average token logits & (4) & \cmark & \cmark & \xmark & classification; aspect term extraction \\
\midrule
\cite{wei-etal-2021-shot} & i & TA-H & token perturbation strength & (3) & \cmark & \cmark & \cmark & classification; text classification \\
\midrule
\cite{wei-etal-2021-trigger} & i & TD-H & proportion of related arguments in the input & (5) & \cmark & \cmark & \cmark & classification; event argument extraction \\
\midrule
\cite{wenjing-etal-2021-improving} & g & TD-M & average model confidence & (4) & \cmark & \cmark & \cmark & classification; NER \\
\midrule
\cite{agrawal-carpuat-2022-imitation} & g & TD-H & Levenshtein distance & (4) & \cmark & \cmark & \cmark & seq2seq; simplification \\
\midrule
\cite{chen-etal-2022-crossroads} & g & TA-H & maximum entity length + entity complexity & (4) & \cmark & \cmark & \cmark & classification; NER \\
\midrule
\cite{christopoulou-etal-2022-training} & g & TD-M & variability; correctness; confidence & (2) & \cmark & \cmark & \xmark & classification; NLU; multilingual NLU \\
\midrule
\cite{christopoulou-etal-2022-training} & g & TD-M & variability; correctness; confidence & (4) & \cmark & \cmark & \xmark & classification; NLU; multilingual NLU \\
\midrule
\cite{grishina-sorokin-2022-local} & i & TD-H & label-masking ratio & (3) & \cmark & \cmark & \cmark & classification; intent classification \\
\midrule
\cite{lee-etal-2022-efficient-pre} & i & TA-H & degree of connection & (2) & \xmark & \xmark & \xmark & pre-training \\
\midrule
\cite{li-etal-2022-complex} & i & TA-H & length of KG sequences & (3) & \cmark & \cmark & \cmark & reasoning; temporal KG extrapolation \\
\midrule
\cite{li-etal-2022-curriculum} & g & TA-H & degree of prompt perturbation & (3) & \cmark & \cmark & \cmark & generation; dialogue summarisation \\
\midrule
\cite{lobov-etal-2022-applying} & g & TA-H & sentence length & -- & -- & -- & -- & classification; NER \\
\midrule
\cite{lobov-etal-2022-applying} & g & TA-M & perplexity & -- & -- & -- & -- & classification; NER \\
\midrule
\cite{lobov-etal-2022-applying} & g & TD-M & classifier confidence & -- & -- & -- & -- & classification; NER \\
\midrule
\cite{maharana-bansal-2022-curriculum} & l & TD-M & energy & (1) & \xmark & \xmark & \xmark & classification; NLU \\
\midrule
\cite{maharana-bansal-2022-curriculum} & l & TD-M & variability & (1) & \xmark & \xmark & \xmark & classification; NLU \\
\midrule
\cite{maharana-bansal-2022-curriculum} & l & TD-M & question answering probability & (1) & \xmark & \xmark & \xmark & classification; NLU \\
\midrule
\cite{vakil-amiri-2022-generic} & l & TD-M & Trend-SL & (5) & \xmark & \xmark & \xmark & classification; relation extraction \\
\midrule
\cite{wang-etal-2022-feeding} & g & TA-H+ TA-M+ TD-H & hybrid linguistic complexity + semantic similarity + reference-based features & (4) & \cmark & \cmark & \cmark & generation; question answering \\
\midrule
\cite{wang-etal-2022-hierarchical} & g & TD-H & AMR graph depth & (3) & \xmark & \xmark & \xmark & seq2seq; abstract meaning representation parsing \\
\midrule
\cite{yang-etal-2022-take} & g & TA-H & entity-in-context heuristic (distant supervision) & (5) & \cmark & \xmark & \cmark & generation; dialogue system \\
\midrule
\cite{yang-song-2022-fpc} & g & TD-H & inclusion of target tokens & (5) & \cmark & \cmark & \cmark & classification; relation extraction \\
\midrule
\cite{yuan-etal-2022-generative-entity} & l & TA-H+ TD-M & type length + data source + model loss & (4) & \cmark & \cmark & \cmark & classification; generative entity typing \\
\midrule
\cite{zheng-etal-2022-distilling} & l & TD-M & model confidence & (4) & \cmark & \cmark & \cmark & classification; NER \\
\midrule
\cite{elbayad-etal-2023-fixing} & g & TA-H & per-language sample count & (4) & \cmark & \cmark & \cmark & seq2seq; multilingual machine translation \\
\midrule
\cite{elbayad-etal-2023-fixing} & g & TD-M & per-language overfitting onset step & (4) & \cmark & \cmark & \cmark & seq2seq; multilingual machine translation \\
\midrule
\cite{elgaar-amiri-2023-hucurl} & g & TD-H & annotation entropy & (5) & \xmark & \xmark & \xmark & classification \\
\midrule
\cite{elgaar-amiri-2023-hucurl} & g & TD-M & loss & (5) & \xmark & \xmark & \xmark & classification \\
\midrule
\cite{elgaar-amiri-2023-ling} & l & TA-H+ TD-M & model-weighted linguistic features & (5) & \cmark & \xmark & \cmark & classification; NLU \\
\midrule
\cite{jia-etal-2023-sample} & i & TD-H & prefix length of target output & (3) & \cmark & \cmark & \cmark & generation; question answering \\
\midrule
\cite{kakkar-etal-2023-search} & i & TA-H & spelling error complexity class & (3) & \xmark & \xmark & \xmark & seq2seq; spelling correction \\
\midrule
\cite{liu-etal-2023-code} & g & TA-H+ TD-M & program complexity + model loss & (4) & \cmark & \cmark & \cmark & continued pre-training; code execution \\
\midrule
\cite{lu-lam-2023-pcc} & g & TA-M & textual similarity score (BLEURT) & (3) & \xmark & \xmark & \xmark & classification and dialogue generation \\
\midrule
\cite{vakil-amiri-2023-curriculum} & g & TD-H & 26 graph complexity features with regards to label & (3) & \cmark & \cmark & \cmark & classification; node classification and link prediction \\
\midrule
\cite{wu-etal-2023-empower} & i & TA-H & number of nested boolean operations & (4) & \cmark & \cmark & \cmark & reasoning; classification; question answering \\
\midrule
\cite{dou-etal-2024-stepcoder} & i & TD-H & number of conditional statements in the target & (3) & \cmark & \cmark & \cmark & generation; code generation \\
\midrule
\cite{hamad-etal-2024-fire} & g & TD-H & human labelling effort proxy & (4) & \cmark & \cmark & \cmark & classification; relation extraction; information extraction \\
\midrule
\cite{jia-li-2024-metaphor} & g & TD-M & cross-entropy loss & (4) & \cmark & \cmark & \cmark & classification; metaphor detection \\
\midrule
\cite{jin-etal-2024-enhancing} & i & TD-M & decision boundary proximity & -- & -- & -- & -- & classification \\
\midrule
\cite{li-etal-2024-active} & g & TA-M & LLM (GPT3.5) rated difficulty & (4) & \cmark & \cmark & \xmark & seq2seq; summarisation \\
\midrule
\cite{liu-etal-2024-curriculum} & l & TD-M & consistency learning loss & (5) & \xmark & \xmark & \xmark & seq2seq; speech-to-text; multimodal machine translation \\
\midrule
\cite{liu-etal-2024-fisher} & g & TD-M & Fisher Information Matrix trace & (4) & \cmark & \cmark & \cmark & classification; question answering \\
\midrule
\cite{mohankumar-etal-2024-improving} & i & TA-H & context length & (3) & \cmark & \cmark & \cmark & key word retrieval \\
\midrule
\cite{pattnaik-etal-2024-enhancing} & g & TD-H & human rating difference & (3) & \cmark & \cmark & \cmark & alignment; direct preference optimization \\
\midrule
\cite{pattnaik-etal-2024-enhancing} & g & TD-M & GPT4 rating difference & (3) & \cmark & \cmark & \cmark & alignment; direct preference optimization \\
\midrule
\cite{pattnaik-etal-2024-enhancing} & l & TD-M & model rating gap between chosen and rejected answers & (5) & \cmark & \cmark & \cmark & alignment; direct preference optimization \\
\midrule
\cite{poesina-etal-2024-novel} & g & TD-M & training dynamics signals (confidence and variability) & (3) & \cmark & \cmark & \cmark & classification; NLU \\
\midrule
\cite{ranaldi-etal-2024-language} & g & TA-H+ TD-M & level of Bloom’s taxonomy + input length & (3) & \cmark & \cmark & \cmark & post-training; instruction tuning \\
\midrule
\cite{wang-etal-2024-theoremllama} & g & TD-H & proof step count & -- & -- & -- & -- & generation; theorem proving \\
\midrule
\cite{weinzierl-harabagiu-2024-discovering} & g & TA-M & euclidean distance between SBERT embeddings & (4) & \cmark & \cmark & \cmark & in-context learning; generative framing discovery \\
\midrule
\cite{wu-etal-2024-instruction} & g & TA-H & pre-defined levels of difficulty & (3) & \cmark & \cmark & \cmark & generation; code generation \\
\midrule
\cite{xu-etal-2024-automatic} & g & TA-M & performance gap between generator models & (5) & \cmark & \cmark & \cmark & post-training; alignment \\
\midrule
\cite{yu-etal-2024-popalm} & l & TD-M & ROUGE-based reward enhancement + reward-model confidence & (1) & \xmark & \xmark & \xmark & generation; trendy social media responses \\
\midrule
\cite{brei-etal-2025-classifying} & g & TD-M & count of plausible label candidates as identified by an LLM & (3) & \cmark & \cmark & \cmark & classification \\
\midrule
\cite{elhady-etal-2025-emergent} & g & TA-H & binary language attribute & (2) & \cmark & \cmark & \cmark & continued pre-training; language adaptation \\
\midrule
\cite{elzohbi-zhao-2025-tahdib} & g & TA-H & pre-defined domain complexity & (3) & \cmark & \cmark & \cmark & generation; arabic poetry editing \\
\midrule
\cite{hong-etal-2025-game} & g & TA-H & pre-defined levels of difficulty & (3) & \cmark & \cmark & \cmark & generation; code generation \\
\midrule
\cite{pan-etal-2025-negative} & g & TA-M & cosine similarity between queries and negative samples & (3) & \cmark & \cmark & \cmark & semantic textual similarity; retrieval \\
\midrule
\cite{ren-etal-2025-step} & g & TA-H & number of constrains & (2) & \cmark & \cmark & \cmark & post-training; instruction tuning \\
\midrule
\cite{tang-etal-2025-effective} & l & TD-M & Bayesian Uncertainty & (4) & \xmark & \cmark & \xmark & classification; sequence labelling \\
\midrule
\cite{toborek-etal-2025-beyond} & g & TA-H & human-curated register labels & (3) & \cmark & \cmark & \cmark & pre-training \\
\midrule
\cite{wang-etal-2025-knowledge} & i & TD-M & noise level & (3) & \cmark & \cmark & \cmark & classification; KG entity typing \\
\midrule
\cite{yao-etal-2025-language-models} & g & TA-H & hop-count (number of reasoning steps) & (4) & \cmark & \cmark & \cmark & generation; reasoning \\
\midrule
\cite{zhang-etal-2025-learning-like} & l & TD-M & model success rate & (3) & \xmark & \xmark & \xmark & post-training; mathematical reasoning \\
\midrule
\cite{zhang-etal-2025-preference} & g & TD-M & perplexity difference (PD) between strong and weak model & (5) & \cmark & \cmark & \cmark & pre-training \\
\end{longtable}
\endgroup

\end{document}